%% file: main.tex
\documentclass[11pt,twocolumn,letterpaper]{article}
\usepackage[left=0.8in,right=0.8in,top=0.8in,bottom=0.8in]{geometry}

\usepackage{times}
\usepackage{epsfig}
\usepackage{graphicx}
\usepackage{amsmath,amssymb,bm}
\usepackage{multirow}

\usepackage{booktabs}
\usepackage{blindtext}
\usepackage{hyperref}
\usepackage{comment}
\usepackage{subcaption}
\usepackage{xspace}
\usepackage{colortbl}

\usepackage[capitalize]{cleveref}
\crefname{section}{Sec.}{Secs.}
\Crefname{section}{Section}{Sections}
\Crefname{table}{Table}{Tables}
\crefname{table}{Tab.}{Tabs.}

\newcommand{\ie}{\textit{i.e.,}}
\newcommand{\etal}{\textit{et al.}}

\definecolor{mentor-green}{RGB}{0,170,70}
\definecolor{mentor-blue}{RGB}{60,100,190}

\begin{document}

\title{Grains of Saliency:\\Optimizing Saliency-based Training of Biometric\\ Attack Detection Models}

\author{Colton R. Crum\hspace{2cm}Samuel Webster\hspace{2cm}Adam Czajka\\
Department of Computer Science and Engineering \\
University of Notre Dame, IN, USA\\
{\tt\small \{ccrum,swebster,aczajka\}@nd.edu}}
\date{}

\maketitle

\begin{abstract}
Incorporating human-perceptual intelligence into model training has shown to increase the generalization capability of models in several difficult biometric tasks, such as presentation attack detection (PAD) and detection of synthetic samples. After the initial collection phase, human visual saliency (e.g., eye-tracking data, or handwritten annotations) can be integrated into model training through attention mechanisms, augmented training samples, or through human perception-related components of loss functions. Despite their successes, a vital, but seemingly neglected, aspect of any saliency-based training is the level of salience granularity (e.g., bounding boxes, single saliency maps, or saliency aggregated from multiple subjects) necessary to find a balance between reaping the full benefits of human saliency and the cost of its collection. In this paper, we explore several different levels of salience granularity and demonstrate that increased generalization capabilities of PAD and synthetic face detection can be achieved by using simple yet effective saliency post-processing techniques across several different CNNs.
\end{abstract}

\section{Introduction}
\label{sec:ijcb-introduction}
An ongoing challenge across biometric presentation attack detection (PAD) involves obtaining sufficient model generalization to unknown attack types, or unknown variants of known attacks. For arduous tasks such as PAD, model generalization is crucial as new attack types show up frequently and cannot be meaningfully represented with sufficient training samples, if at all. Despite the accelerated gains made in recent years for closed-set recognition tasks (all attack types, or their variants known during training), these trends have failed to fully materialize in biometrics due to their appetite for mammoth amounts of training data. As a consequence, state-of-the-art (SOTA) biometric PAD performance is lackluster at best, as evidence from the recurrent Liveness Detection (LivDet) competitions for face \cite{purnapatra2021face}, iris \cite{yambay2017livdet, das2020iris, tinsley2023iris}, and fingerprint \cite{yambay2018livdet, orru2019livdet, casula2021livdet}.
The incompetence of models on attack types trivial to a human (e.g., doll eye for iris) can be mitigated by incorporating human-perceptual intelligence into model training through attention mechanisms \cite{linsley2018learning}, human saliency-guided data augmentations \cite{boyd2022human}, or loss function components penalizing divergence of model's saliency from human saliency \cite{boyd2021cyborg}. Despite their success, human saliency-based methods are significantly diminished by the sheer costs associated with acquisition (monetary, time, availability of human subjects), often curtailing its practical use.
However, a seemingly obvious but universally unanswered question at the cornerstone of all saliency-based methods is the level of detail, or granularity, necessary to achieve the desired generalization. Salience granularity can be explored through the initial acquisition phase (e.g., high-resolution eye fixation, or simply bounding boxes roughly approximating human salience), or through post-processing measures (e.g., averaging annotation maps from multiple experts, or using a single expert). Understanding the optimal level of salience granularity leads to a better assessment of the costs associated with salience acquisition. In addition, salience granularity validates the ability of saliency-based methods to incorporate salient information meaningfully into model training. Without this necessary foundation, how the collection and assembly of saliency impacts the generalization performance of saliency-based training downstream remains unclear. We find that for iris PAD and synthetic face detection, single saliency maps provide a sufficient amount of human-sourced information, which suggests that the collection of fine-grained salience may not be necessary. Additionally, it answers the previous unrealized trade-off between the quality (granularity) and quantity of human salient-information necessary for successful saliency-based training. Our results suggest that the quantity of saliency contributes more to model generalization more than its quality (depending on the biometric modality). Furthermore, we explore salience granularity across several different sources, including human subjects, models trained to mimic human saliency, and domain-specific segmentation models. We find that substantial performance gains can be made within saliency-based training by using optimal salience granularity with no additional overhead.

We organized our paper around the following research questions:
\begin{itemize}
    \itemsep0em
    \item \textbf{RQ1}: What is the optimal level of granularity of human saliency maps for saliency-based training of models detecting biometric spoofs?
    \item \textbf{RQ2}: Does the optimal level of granularity generalize across different biometric presentation attack instruments?
    \item \textbf{RQ3}: Do models trained to mimic human saliency offer image annotations which -- when used in saliency-guided trained -- lead to better generalization?
    \item \textbf{RQ4}: Can saliency be sourced from domain-specific segmentation models instead of humans?
\end{itemize}

We release our source codes, model weights and saliency maps to allow others to replicate all the experiments.\footnote{Source codes will be released when the peer-reviewed version is published.}

\section{Related Work}
\label{sec:ijcb-related-work}

Incorporating human-perception into the training of deep learning models has shown to increase generalization performance \cite{boyd2021cyborg, boyd2020iris, wang2024gazegnn}, create more human-interpretable outputs \cite{crum2023explain, wang2024gazegnn}, increase overall training efficiency by requiring less samples \cite{van2023probabilistic}, and even help ease regulatory tensions within high-risk AI \cite{crum2023seriously}. Domains such as medical imaging and biometrics largely benefit from incorporating human-salient information into training since these scenarios pose challenging constraints not common within general vision tasks. For biometric presentation attack detection, models often operate within open-set contexts where it is unrealistic or otherwise impossible to obtain training samples for every possible attack type. As a result, human-perceptual components have proven invaluable towards increasing generalization capabilities by augmenting training samples \cite{boyd2020iris}, attention modules \cite{linsley2018learning, wang2024gazegnn}, or through loss function components \cite{boyd2021cyborg, piland2023model, van2023probabilistic}.

While methods on how to effectively incorporate human saliency into model training are important, arguably a more crucial aspect of saliency-based training is how to effectively source and assembly the raw saliency, which is often overlooked. Many saliency-based rely upon eye-tracking \cite{van2023probabilistic}, gaze patterns \cite{wang2024gazegnn}, or written annotations \cite{boyd2021cyborg, boyd2020iris}. After the initial collection phase, most post-processing methods include averaging correctly classified samples together into a single saliency map. However, this vastly reduces the number of available saliency and leaves unanswered questions as to how this might affect models trained downstream.

\section{Methodology}
\label{sec:ijcb-methods}
\input{figures/all-saliency}

In this section, we first describe the training, validation, and testing datasets. Second, we define three levels of salience granularity and describe several sources of salience used within saliency-based training. Finally, we describe the training and evaluation procedure used to evaluate the research questions presented in the Introduction.

\subsection{Datasets}
\label{sec:ijcb-methods-datasets}
We evaluate the affect of salience granularity from several sources for iris-PAD and synthetic face detection tasks. These tasks were selected due to their availability of human annotations necessary to explore granularity across consistent amounts of salient-information, and have proven useful using the CYBORG loss function \cite{boyd2021cyborg}.

\paragraph{Training \& Validation Set} For the \textbf{iris-PAD} task, training and validation images were sampled from a superset composed of various live iris and iris PAD datasets \cite{casia-database,Sung_OE_2007,Galbally_ICB_2012,Kohli_ICB_2013,Yambay_ISBA_2017,Trokielewicz_IVC_2020,Kohli_BTAS_2016,Wei_ICPR_2008,Trokielewicz_BTAS_2015,Yambay_IJCB_2017,Das_IJCB_2020}. The training set consisted of 765 samples comprising of bona fide (live) and seven spoof attack types (artificial, diseased, post mortem, paper print outs, synthetic, textured contact lens, textured contact lens \& printed), offered by \cite{boyd2020iris}. The validation set comprised of 23,312 samples, completely disjoint from training and testing sets.

For the \textbf{synthetic face detection} task, we follow the training splits introduced in \cite{boyd2021cyborg}\footnote{The authors of this paper would like to thank the authors of \cite{boyd2021cyborg} for sharing their data and training splits with us.}. The training set consisted of 1821 (919 real and 902 synthetic), and the validation set consisted of 20,000 samples (10,000 real and 10,0000 synthetic) extracted from the FRGC-Subset~\cite{Phillips_IVC_2017}, SREFI \cite{Banerjee_IJCB_2017} and StyleGAN2-generated acquired from \emph{thispersondoesnotexist.com}.

\paragraph{Test Set} For the \textbf{iris-PAD} task, we use LivDet-2020 to benchmark model generalization performance \cite{das2020iris}. This edition of LivDet is particularly useful within saliency-based training as a variety of attack types can be meaningfully annotated by human subjects (\ie post mortem), which translate downstream to raise generalization performance. Our paper aims to explore the optimal salience granularity (by first using configurations that have already proven to work), and not introduce attack-type specifics that distract from our analysis.

For the \textbf{synthetic face detection task}, we sub-sampled the test set provided by \cite{boyd2021cyborg} to reduce computational overhead. 500 images were randomly without replacement from two live (FFHQ \cite{karras2017progressive} and CelebA-HQ \cite{karras2017progressive}) and 1000 images from six synthetic, GAN-generated sources (ProGAN \cite{karras2017progressive}, StarGANv2 \cite{stargan}, StyleGAN \cite{karras2017progressive}, StyleGAN2 \cite{StyleGAN2}, StyleGAN2-ADA \cite{Karras2020ada}, and StyleGAN3 \cite{karras2021sg3}. In total, the synthetic face detection test set comprised of 7,000 (1000 live and 6000 synthetic) test images.

\subsection{Acquisition of Salience}
In this section, we first define three levels of salience granularity (Boundary of Interest, Area of Interest, and Features of Interest), aimed to provide a varied amount of salience. Next, we describe three sources of salience (human subjects, models trained to mimic human subjects, and domain-specific segmentation models). These configurations allow for a cross-sectional analysis surrounding the optimal salience granularity and from which particular source. For our experiments, we use the human saliency provided by previous works for iris-PAD and synthetic face detection \cite{boyd2021cyborg, boyd2020iris}.

\paragraph{Granularity of Human Salience}
As described in Sec.\ref{sec:ijcb-related-work}, previous work spent little time exploring effective means of assembling salience from human annotations. Most works simply averaged annotations from correctly classified samples, aggregating these samples into a single saliency map \cite{boyd2020iris, boyd2021cyborg}. As a consequence, the number of training samples with accompanying saliency maps was pruned to a meager few hundred samples. In Boyd \etal, 10,750 saliency maps collected were reduced to only 1,821 \cite{boyd2021cyborg}. This configuration prioritizes fine-grained, human-salient features at the expense of an abundant supply. However, to our knowledge, no prior work has viewed human saliency under a variable approach, or within different levels of granularity.

In this work, we evaluate salience (sourced by human subjects and models trained to mimic human subjects) under the following levels of granularity (see Fig. \ref{fig:ijcb-saliency}):
\begin{itemize}
    \itemsep0em
    \item \textbf{Boundary of Interest (BOI)}: a rectangular box that indicates a general boundary pertaining to human saliency.
    \item \textbf{Area of Interest (AOI)}: a binarized region that indicates human saliency \emph{uniformly}.
    \item \textbf{Features of Interest (FOI)}: regions that indicate fine-grained, \emph{variable} amounts human saliency within specific regions of the input sample.
\end{itemize}

Since previous works average saliency maps together to obtain fine-grained detail within each sample, we describe this as Features of Interest (FOI). This assembly prioritizes sample-level features pertinent towards solving the larger task. In an effort to standardize the amount of salient-information contained within each sample while maintaining a varied degree of granularity, Boundary of Interest (BOI) and Area of Interest (AOI) salience was derived from the FOI salience. AOI salience was generated by first binarizing FOI salience, wherein pixels with values greater than 0 were set to 255. For BOI salience, a minimally enclosing rectangle was drawn that encompasses all salient regions found within the FOI salience. The same process was conducted from salience generated by both human subjects and models trained to mimic human subjects (described in the section below).

\paragraph{Models Mimicking Human Subjects} Given the constraints associated with collecting salience from human subjects, a more efficient use of human saliency is to train a model to generate or emulate human saliency. Under this scenario, an autoecoder-type model is trained to learn human-salient features given a cooresponding input image. Once trained, the model can generate salience easily at scale (unlike human saliency). To explore the potential sources of salience, autoencoders were trained using human-sourced (FOI) saliency maps (described above) and used to generate salience for a second training split, comprised of the same size (764 samples for iris-PAD, 1821 samples for synthetic face) and sources as detailed above. AOI and BOI saliency were derived from the FOI saliency sourced by the autoencoder for both respective tasks. For iris-PAD, AOI saliency was derived by binarizing the FOI saliency with a threshold of 0.5, wherein pixels with values greater than 127 were set to 255 and lesser or equal values were set to 0. BOI saliency was drawn with a minimally enclosing rectangle over the salient regions found within the AOI saliency eroded by a \(3\times3\) kernel, initialized uniformly at 1.0 for a single iteration. For the synthetic face detection task, AOI saliency was likewise derived by binarizing the FOI saliency with a threshold of 0.5. BOI saliency was also generated from the AOI saliency, except no erosion was applied prior to computation. For convenience in training the autoencoders, a lightweight parameter sweep was performed using RayTune for efficient traning of the autoencoder for each respective task\cite{liaw2018tune}. For iris-PAD, a DenseNet-161-based UNET \cite{huang2017densely, ronneberger2015u} was trained using the Adam optimizer ($lr=0.0001$) for 50 epochs and a batch size of 20. For the synthetic face detection task, an Inception-V4-based UNET \cite{szegedy2016inceptionv4, ronneberger2015u} was trained using the Adam optimizer ($lr=0.0001$) for 50 epochs and a batch size of 10. Both models were trained using a sigmoid activation function, initialized with ImageNet weights \cite{ImageNet}, and using a Mean Squared Error (MSE) loss, sourced from \cite{Iakubovskii:2019}.

\paragraph{Segmentation Models} Domain-specific segmentation models offer feature-level masks that may be useful in saliency-based training, requiring no overhead and essentially for free. Additionally, if domain-specific segmentation models can achieve comparable performance to human subjects, human salience collections may be antiquated and completely unnecessary. We explore the validity of SOTA segmentation-sourced saliency for iris-PAD using CC-NET \cite{mishra2021image, czajka2023irisrepo}, and the BiSeNet-based face parser \cite{zllrunning2019parse, yu2018bisenet} for synthetic face. Models were evaluated off the shelf (without any fine-tuning or training), and segmentations were generated using the original training split described in Sec.\ref{sec:ijcb-methods-datasets} (see column ``Seg'' in Fig. \ref{fig:ijcb-saliency}).

\subsection{Training \& Evaluation Configurations}
The various configurations of salience granularity and salience sources was evaluated across three CNN architectures, ResNet50 \cite{he2016deep}, DenseNet-121 \cite{huang2017densely}, and Inception-V3 \cite{szegedy2016rethinking} using the CYBORG loss function \cite{boyd2021cyborg}. The CYBORG loss parameter weighting human-guidance and classification performance was given equal weight ($\alpha=0.5$) for all salience granularity experiments (BOI, AOI, and FOI), as in \cite{boyd2021cyborg}. Baseline training without the use of any salience was performed with cross-entropy loss (``None'' in Tab. \ref{tab:ijcb-iris-results} and Tab. \ref{tab:ijcb-face-results}) under the same training configurations for additional comparison. All models were initialized with ImageNet weights \cite{ImageNet}, trained with a batch size of 20 using Stochastic Gradient Descent (SGD) for 50 epochs, with learning rate of 0.005, modified by a scheduler, which reduced the learning rate by a factor of 0.1 every 12 epochs.
We evaluate the generalization performance on the test set using Area Under the ROC Curve (AUC). For the \textbf{iris-PAD task}, the means and standard deviations of AUC are reported across 3 independent runs in Tab. \ref{tab:ijcb-iris-results}, whereas the \textbf{synthetic face detection} models are reported across 5 independent runs (Tab. \ref{tab:ijcb-face-results}). Since our analysis is focused solely on measuring granularity, we do not include generalization results from previous work as they use the same granularity of salience for all experiments (FOI granularity).

\section{Results}
\label{sec:ijcb-results}

This section describes the results, organized by the research questions presented in the Introduction. Results for all variants are summarized in Tab. \ref{tab:ijcb-iris-results} and Tab. \ref{tab:ijcb-face-results}, and ROC curves are displayed in Fig. \ref{fig:results-AUC-iris} and Fig. \ref{fig:results-AUC-face}.

\subsection{RQ1: What is the optimal level of granularity?}
\label{sec:ijcb-RQ1}
\input{tables/all-iris-results}

Focusing first on salience sourced from human subjects, granularity has a sizable impact on generalization performance for iris-PAD (see Tab. \ref{tab:ijcb-iris-results}). The conventional configuration of salience, FOI, common previous works, translated downstream to sub-optimal model generalization (average AUC=0.898), compared to AOI (average AUC=0.910) across all three architectures. More specifically, DenseNet had the largest improvement over the conventional saliency selection. These results suggest that strong generalization can be achieved through simpler salience granularity indicating merely Areas of Interest (AOI), which significantly reduces the overhead associated with human subject collection. More specifically, it suggests that detailed saliency methods (\ie averaging saliency maps from multiple annotators, or collecting salience through eye-tracking) does not reap additional generalization benefits for iris-PAD. Furthermore, simply collecting a boundary of salience (BOI in \ref{tab:ijcb-iris-results}) proves insufficient in generalization performance (AUC=0.887), and is on par with using no saliency at all (AUC=0.886). Thus, {\bf the answer to RQ1 is Area of Interest (AOI) salience granularity for iris-PAD}.

\subsection{RQ2: Does optimal granularity generalize across different presentation attack instruments?}
\label{sec:ijcb-RQ2}
\input{tables/all-face-results}

Tab. \ref{tab:ijcb-face-results} indicates that optimal salience granularity generalizes differently for synthetic face detection compared to iris-PAD. More specifically, the results suggest that salience granularity for synthetic face detection tasks are more dependent on the architectures themselves. The optimal salience granularity using human subjects-sourced salience for ResNet architectures was BOI (AUC=0.604), whereas optimal granularity for both DenseNet and Inception architectures was FOI (0.643 and 0.641, respectively). However, it's worth noting the wide standard deviations reported indicate the differences in salience granularity may not be statistically significant for synthetic face detection. These results could also indicate that the human annotations collected for this specific synthetic face detection may not be as valuable as for iris-PAD. However, unlike iris-PAD, including any type of salience during training (BOI, AOI, or FOI) for synthetic face detection on average boosted generalization performance across all architectures (compared to the Baseline, bottom row Tab. \ref{tab:ijcb-face-results}), suggesting that saliency-based gains can be made with quite simple salience over the baseline.

{\bf The answer to RQ2 is negative}: optimal salience granularity is different across biometric modalities (iris-PAD and synthetic face detection), and the use of saliency-based training necessitates thoughtful consideration pertaining to salience granularity and the corresponding architectures used within saliency-based training.

\subsection{RQ3: Do models trained to mimic human saliency offer image annotations which, when used in saliency-guided trained, lead to better generalization?}
\label{sec:ijcb-RQ3}

Arguably the most interesting finding from our experiments is that models trained to mimic human saliency offer impressive gains in generalization performance, including over the human subjects. The majority of model backbones across both tasks (except DenseNet and Inception for synthetic face detection) performed best when models were trained using saliecy sourced from models mimicking human saliency. For iris-PAD, all models held a substantial gains across all granularities (cross model AUC averages include BOI=0.942, AOI=0.960, FOI=0.953) compared to human subjects (BOI=0.887, AOI=0.910, FOI=0.898). Similar to the findings of RQ2, the synthetic face models offered a mixed result of improvement, largely depending on the architecture. Models mimicking human subjects achieved the best average performance out of all salience configurations for ResNet (AOI=0.614), whereas DenseNet and Inception benefited best from fined-grained saliency sourced from human subjects (0.643 and 0.641, respectively). The performance gains using saliency offered by models mimicking human subjects over the human subjects directly can be a result of supplemental salient-information incorporated by the autoencoder (see the differences between BOI, AOI, and FOI between human subjects and models mimicking human subjects in Fig. \ref{fig:ijcb-saliency}). Although the autoencoder was trained to mimic the human annotations, it still has agency in deciding which salient regions to annotate while satisfying the initial human annotation. This process allows an interweaving of complementary human and model salient-information to be encoded directly within the generated saliency map, which can boost model generalization performance downstream with saliency-based training. Finally, these results suggest that the availability of human saliency collections can be expanded without the scaling limitations associated with collection from human subjects. More specifically, these models largely benefited from AOI granularity, which strikes a necessary balance between fine-grained, feature level saliency (FOI) and the unfocused, blanket saliency of BOI.

\subsection{RQ4: Can saliency be sourced from domain-specific
segmentation models instead of humans?}
\label{sec:ijcb-RQ4}

Given our findings in RQ3, a question arises whether saliency collection from human subjects is necessary at all, and whether domain-specific segmentation models can be used instead of human-sourced (\ie by human subjects or models mimicking human subjects). We found that saliency-based training using segmentation models significantly falls short of human-sourced saliency, but is occasionally an improvement from no salience use at all. For iris-PAD, using iris segmenter-based saliency offered the worst average model performance (AUC=0.885), including the baseline with no saliency use at all (AUC=0.886). These results bolster the need for human subjects within the saliency-generation pipeline (either sourced directly from human subjects, or training models to mimic human subjects) for presentation attack detection and related synthetic detection tasks. Domain-specific segmentation models are trained to simply locate basic features of the input sample (\ie annular iris, or nose for faces), which is insufficient information required to solve these tasks. Often the information necessary to correctly classify the PAD sample goes beyond simple feature matching, and requires models to look elsewhere (\ie corners of the sample for post mortem attack types). Unlike domain-specific segmentation masks, human subjects locate these anomalous regions (as do models trained to mimic the human subjects), guiding the models towards salient-regions necessary to solve the PAD task.

{\bf The answer to RQ4 is negative}: saliency is best sourced directly from humans (human subjects or models mimicking human subjects). However, our findings suggest that using saliency from segmentation models may provide generalization gains over traditional configurations where no saliency is used during training.

\section{Conclusion}
\label{sec:ijcb-conclusion}

Efforts to raise generalization in challenging biometric tasks have often incorporated human-perceptual information into the training of CNN models, most commonly through saliency-based training. Despite their ability to improve generalization in iris-PAD and synthetic face detection tasks, obtaining fine-grained salience from human subjects remains an ongoing obstacle. However, our results indicate this challenge may be an illusion for some biometric modalities. In this paper, we find that model generalization can be improved through more manageable collection and assembly. First, we define three levels of salience granularity: Boundary of Interest (BOI), Area of Interest (AOI), and Features of Interest (FOI), which all have varying degrees of detail and associated acquisition costs. Second, we show that traditional salience granularity methods (FOI) is often inferior to more simpler methods (AOI) for iris-PAD tasks. For synthetic face detection, we found that optimal granularity is largely architectural dependent, though models benefited from the use of any level of salience (BOI, AOI, FOI) over no salience use at all during training. Third, we showcase how substantial generalization gains can be made using salience generated by models that mimic human subjects, which combine the complementary information between human subjects and the model. Finally, we show the ineffectiveness of salience sourced from domain-specific models within saliency-based training, encouraging for a human to be involved in the salience curating process. Our paper calls attention to an important, but remarkably missed component to all saliency-based training methods and suggests that generalization performance can be improved through less taxing means of acquisition.

{\small
\bibliographystyle{ieee}
\bibliography{egbib}
}

\input{figures/results-AUC-curves-split}

\end{document}

%% file: figures/all-saliency.tex
\begin{figure*}[!t]
  \centering
  \includegraphics[width=\linewidth]{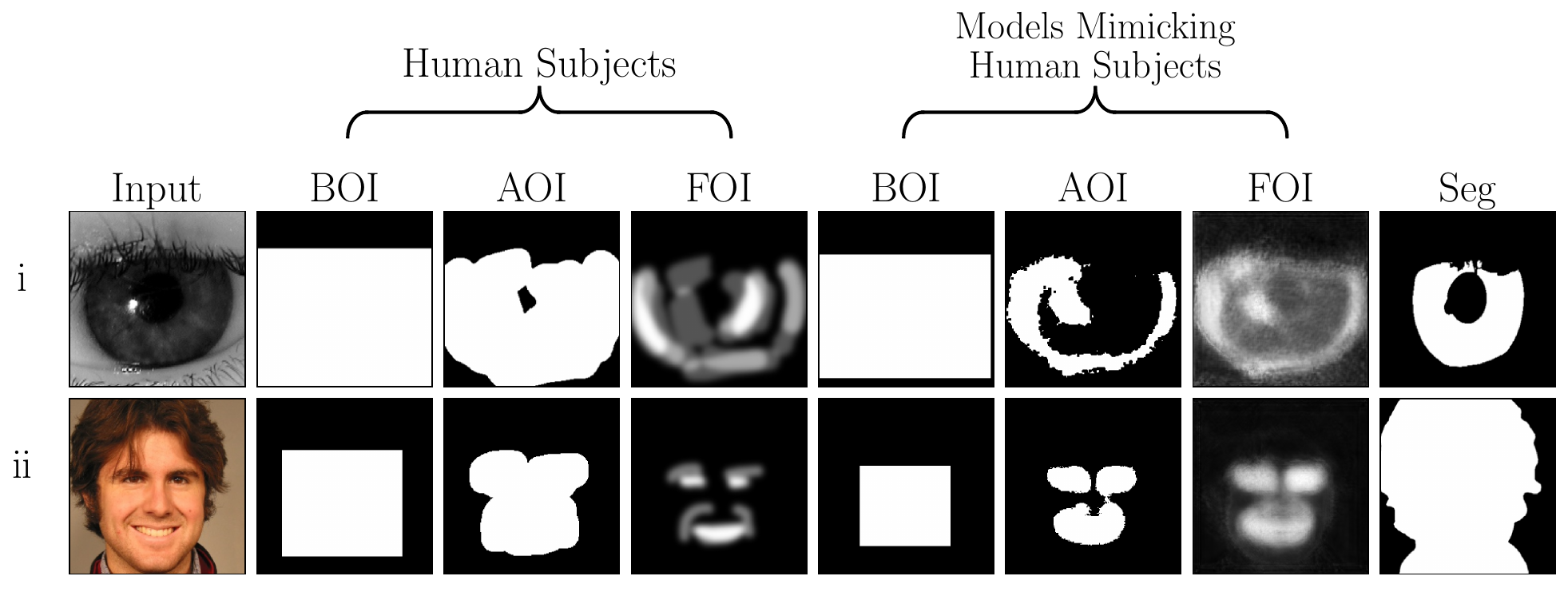}
\caption{Examples of salience granularity used in saliency-based training defined within this paper: Boundary of Interest (BOI), Area of Interest (AOI), Features of Interest (FOI), sourced from either human subjects or models that were trained to mimic the human subjects. ``Seg'' indicates segmentation masks sourced from domain-specific segmentation models; (i) iris presentation attack detection task and (ii) synthetic face detection task.}

\label{fig:ijcb-saliency}
\end{figure*}

%% file: tables/all-iris-results.tex
\begin{table*}[t!]

\centering
\caption{Three CNN architectures (ResNet50, DenseNet-121, and Inception-V3) generalization performance with saliency-based training across various sources of saliency (Human Subjects, Models Mimicking Human Saliency, Segementation Models, or None) across three different salience granularities (only applicable for Human Subjects and Models Mimicking Human Saliency) for \textbf{iris-PAD task}. Means and standard deviations of Area Under the Curve (AUC) are reported across \textbf{3 independent runs}. Optimal saliency configuration for each backbone is \textbf{bolded}, and the optimal average configuration across backbones is \underline{underlined}.}
\label{tab:ijcb-iris-results}
\begin{tabular}{@{}l|lll|ll@{}}
\toprule
\textbf{Source of Saliency} & \textbf{ResNet} & \textbf{DenseNet} & \textbf{Inception} & \textbf{Average} \\
{\bf}  & \multicolumn{3}{c}{Backbones Used in Saliency-Based Training} \\

\midrule
\textbf{Human Subjects} & & & & \\
\hspace{5mm}Boundary of Interest (BOI) & 0.886$\pm$0.015 & 0.903$\pm$0.010 & 0.873$\pm$0.023 & 0.887$\pm$0.016 \\
\hspace{5mm}Area of Interest (AOI) & \text{0.909$\pm$0.006} & \text{0.921$\pm$0.013} & \text{0.900$\pm$0.005} & {0.910$\pm$0.008} \\
\hspace{5mm}Features of Interest (FOI) & \text{0.908$\pm$0.005} & 0.895$\pm$0.018 & 0.890$\pm$0.015 & 0.898$\pm$0.013 \\

\textbf{Models Mimicking Human Subjects} & & & & \\
\hspace{5mm}Boundary of Interest (BOI) & 0.939$\pm$0.008 & 0.933$\pm$0.016 & 0.953$\pm$0.007 & 0.942$\pm$0.010 \\
\hspace{5mm}Area of Interest (AOI) & \textbf{0.956$\pm$0.006} & \textbf{0.962$\pm$0.005} & \textbf{0.962$\pm$0.013} & \underline{0.960$\pm$0.008} \\
\hspace{5mm}Features of Interest (FOI) & \text{0.945$\pm$0.007} & \text{0.955$\pm$0.003} & \text{0.958$\pm$0.007} & 0.953$\pm$0.006 \\

\textbf{Segmentation Models} & & & & \\
\hspace{5mm}Iris Segmentations & 0.894$\pm$0.010 & 0.884$\pm$0.004 & 0.878$\pm$0.022 & 0.885$\pm$0.012 \\

\midrule
\textbf{None} & & & & \\
\hspace{5mm}Baseline & 0.875$\pm$0.013 & 0.893$\pm$0.019 & 0.889$\pm$0.006 & 0.886$\pm$0.010 \\

\bottomrule
\end{tabular}
\end{table*}

%% file: tables/all-face-results.tex
\begin{table*}[t!]

\centering
\caption{Same as Tab. \ref{tab:ijcb-iris-results}, except for \textbf{synthetic face detection task} and results are reported across \textbf{5 independent runs}.}
\label{tab:ijcb-face-results}
\begin{tabular}{@{}l|lll|ll@{}}
\toprule
\textbf{Source of Saliency} & \textbf{ResNet} & \textbf{DenseNet} & \textbf{Inception} & \textbf{Average} \\
{\bf}  & \multicolumn{3}{c}{Backbones Used in Saliency-Based Training} \\

\midrule
\textbf{Human Subjects} & & & & &\\
\hspace{5mm}Boundary of Interest (BOI) & 0.604$\pm$0.048 & 0.546$\pm$0.059 & 0.617$\pm$0.062 & 0.589$\pm$0.056 \\
\hspace{5mm}Area of Interest (AOI) & 0.579$\pm$0.035 & 0.577$\pm$0.045 & 0.639$\pm$0.029 & 0.598$\pm$0.036 \\
\hspace{5mm}Features of Interest (FOI) & 0.590$\pm$0.023 & \textbf{0.643$\pm$0.033} & \textbf{0.641$\pm$0.046} & \underline{0.629$\pm$0.037} \\

\textbf{Models Mimicking Human Subjects} & & & & \\
\hspace{5mm}Boundary of Interest (BOI) & 0.584$\pm$0.031 & 0.583$\pm$0.054 & 0.539$\pm$0.034 & 0.569$\pm$0.040 \\
\hspace{5mm}Area of Interest (AOI) & \textbf{0.614$\pm$0.056} & \text{0.640$\pm$0.046} & 0.608$\pm$0.071 & 0.621$\pm$0.058 \\
\hspace{5mm}Features of Interest (FOI) & 0.600$\pm$0.025 & 0.619$\pm$0.033 & 0.632$\pm$0.019 & 0.617$\pm$0.026 \\

\textbf{Segmentation Models} & & & & \\
\hspace{5mm}Face Segmentations & 0.548$\pm$0.048 & 0.451$\pm$0.050 & 0.579$\pm$0.040 & 0.526$\pm$0.046 \\
\midrule

\textbf{None} & & & & \\
\hspace{5mm}Baseline & 0.572$\pm$0.047 & 0.535$\pm$0.075 & 0.540$\pm$0.037 & 0.549$\pm$0.053 \\

\bottomrule
\end{tabular}
\end{table*}

%% file: figures/results-AUC-curves-split.tex
\begin{figure*}[!t]
  \begin{subfigure}[t]{.33\textwidth}
    \centering
    \includegraphics[width=\linewidth]{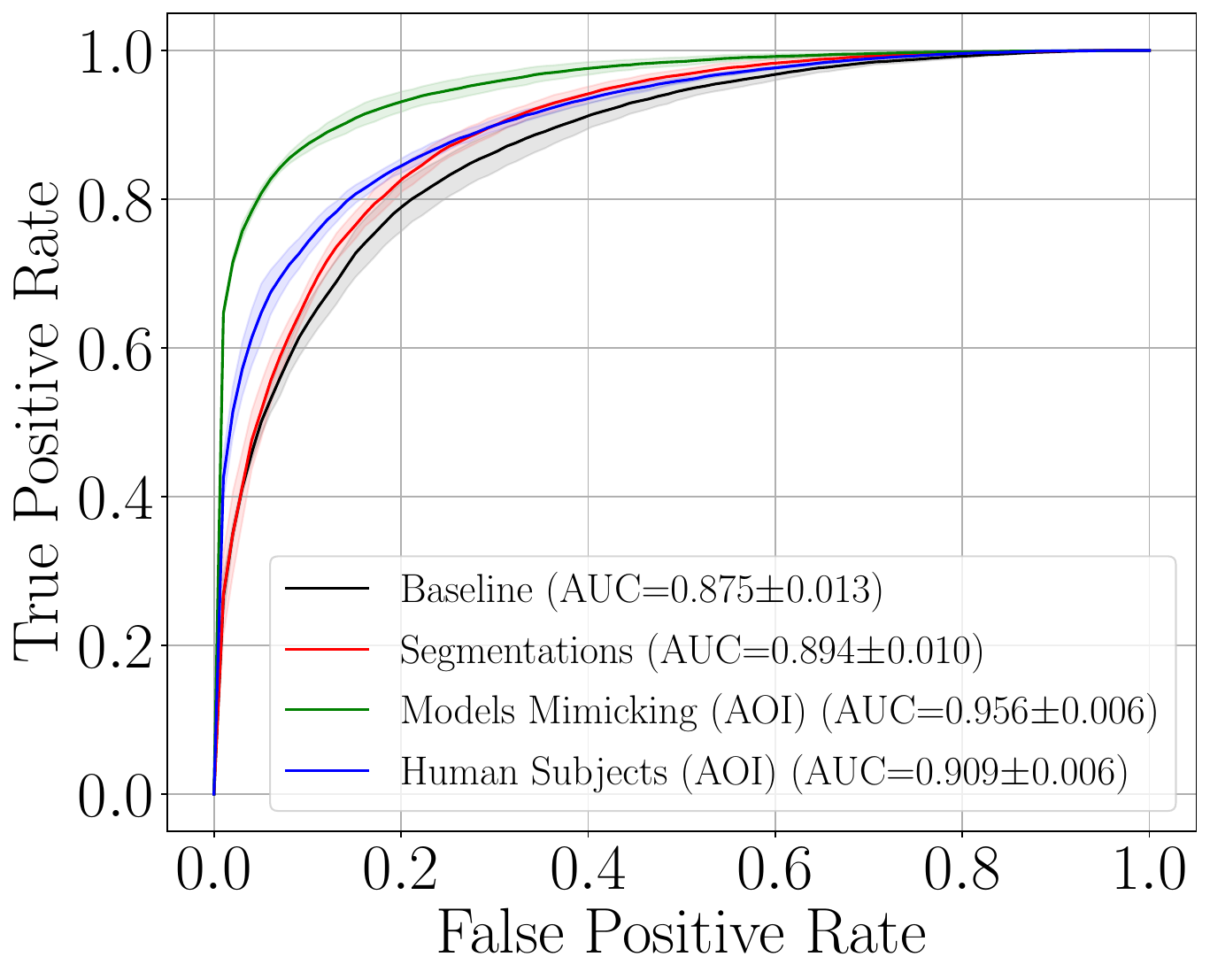}
    \caption{ResNet}
  \end{subfigure}
  \begin{subfigure}[t]{.33\textwidth}
    \centering
    \includegraphics[width=\linewidth]{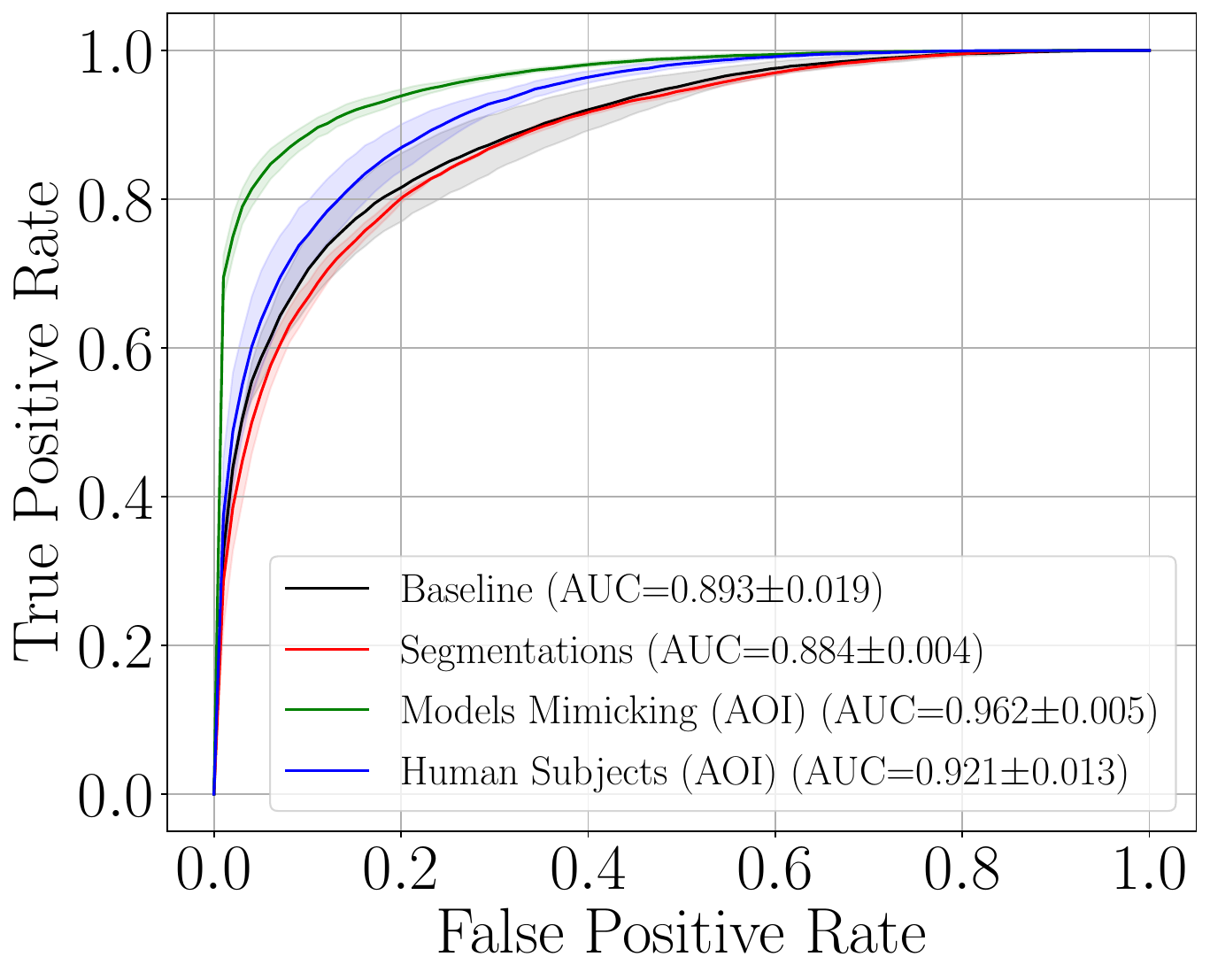}
    \caption{DenseNet}
  \end{subfigure}
  \hfill
  \begin{subfigure}[t]{.33\textwidth}
    \centering
    \includegraphics[width=\linewidth]{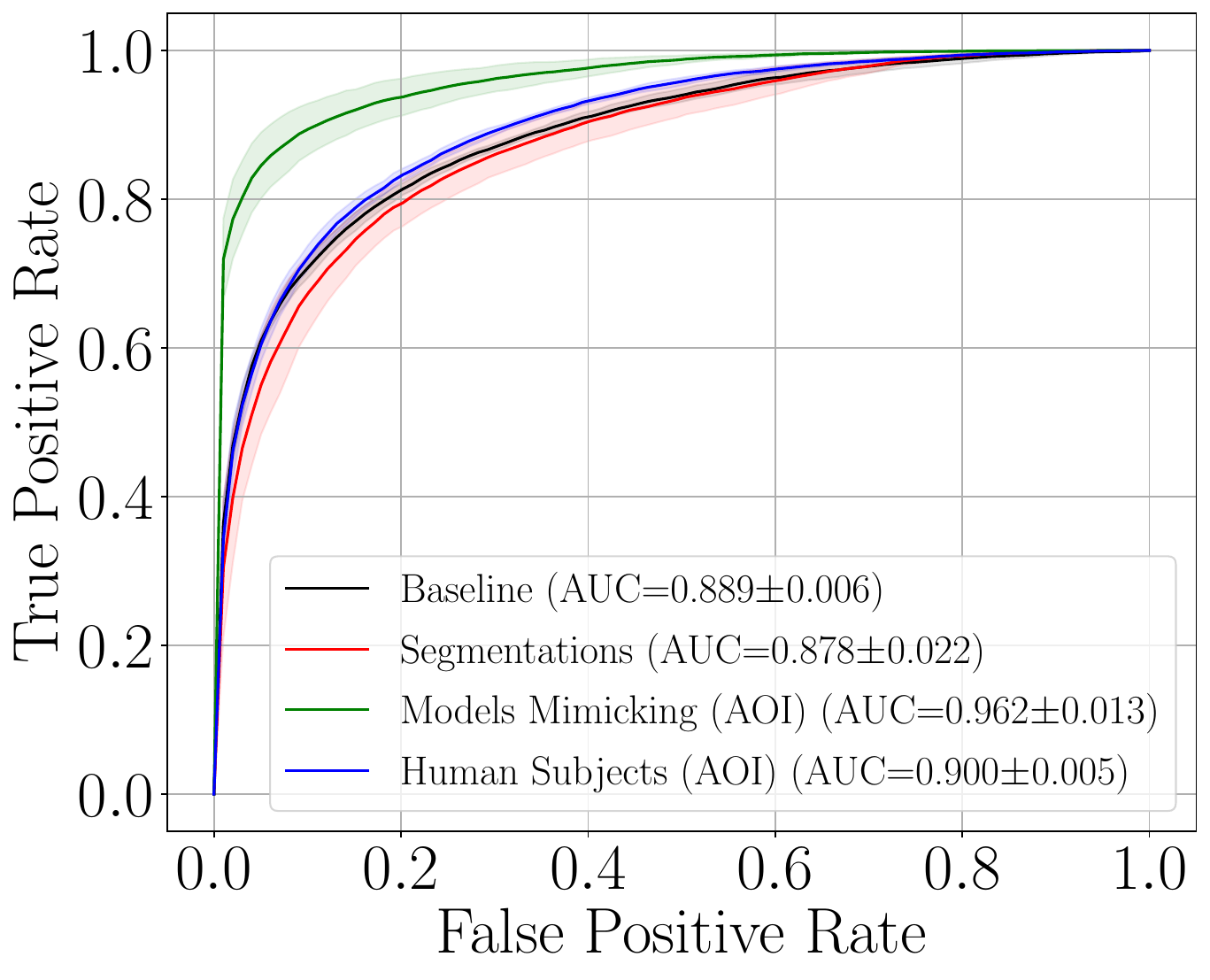}
    \caption{Inception}
  \end{subfigure}
    \caption{Mean ROC curves and bands representing standard deviations (along the True Positive Rate axis) for all backbones used in saliency-based training with varied configurations of saliency for \textbf{iris-PAD (top row)} and \textbf{synthetic face detection (bottom row) tasks}. For human subjects and models mimicking human subjects, the optimal granularity (BBOI, AOI, FOI) is selected, indicating that generalization performance improves having human subjects within the saliency generation pipeline.}
    \label{fig:results-AUC-iris}
\end{figure*}

\begin{figure*}[!t]

    \begin{subfigure}[t]{.33\textwidth}
    \centering
    \includegraphics[width=\linewidth]{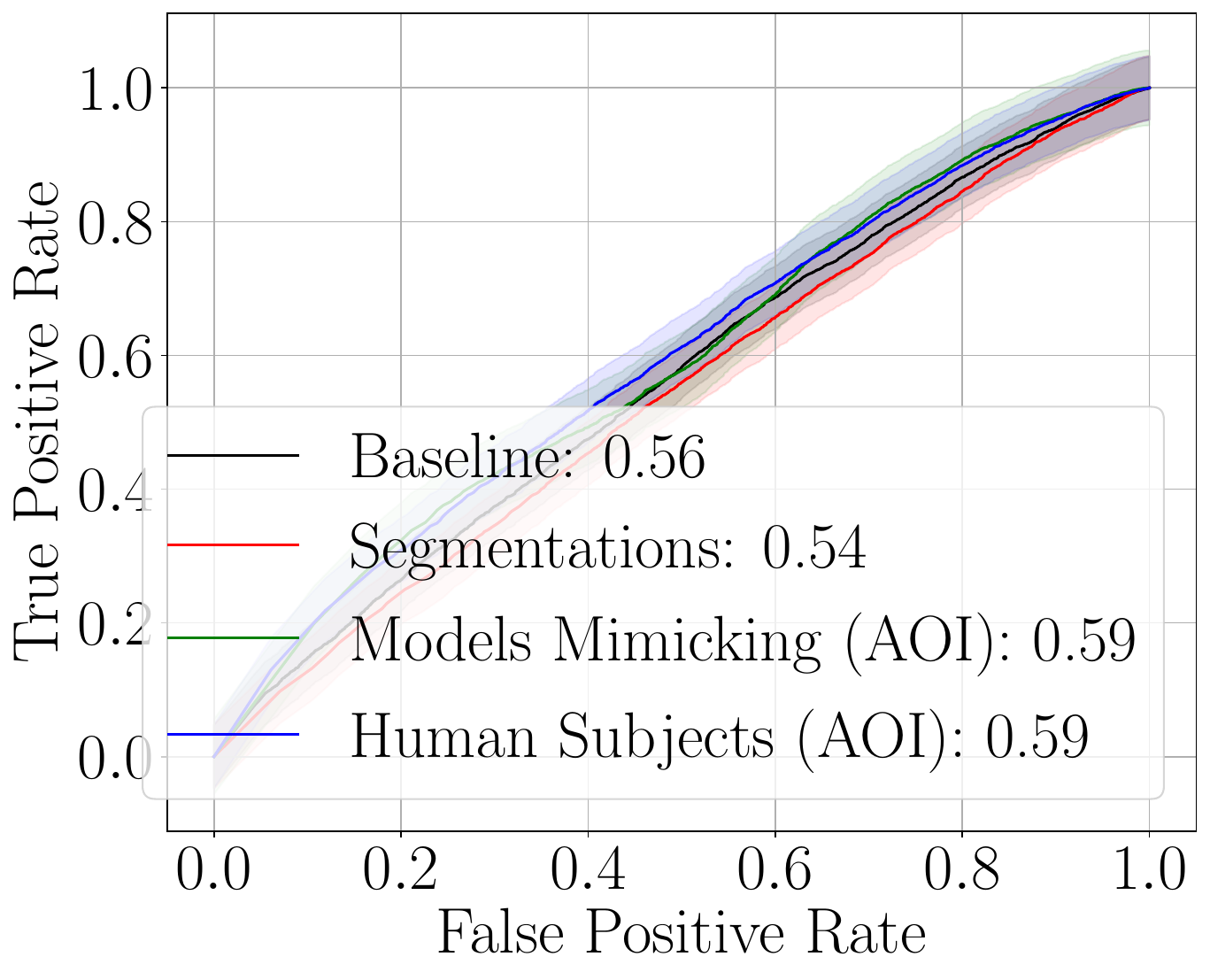}
    \caption{ResNet}
  \end{subfigure}
  \begin{subfigure}[t]{.33\textwidth}
    \centering
    \includegraphics[width=\linewidth]{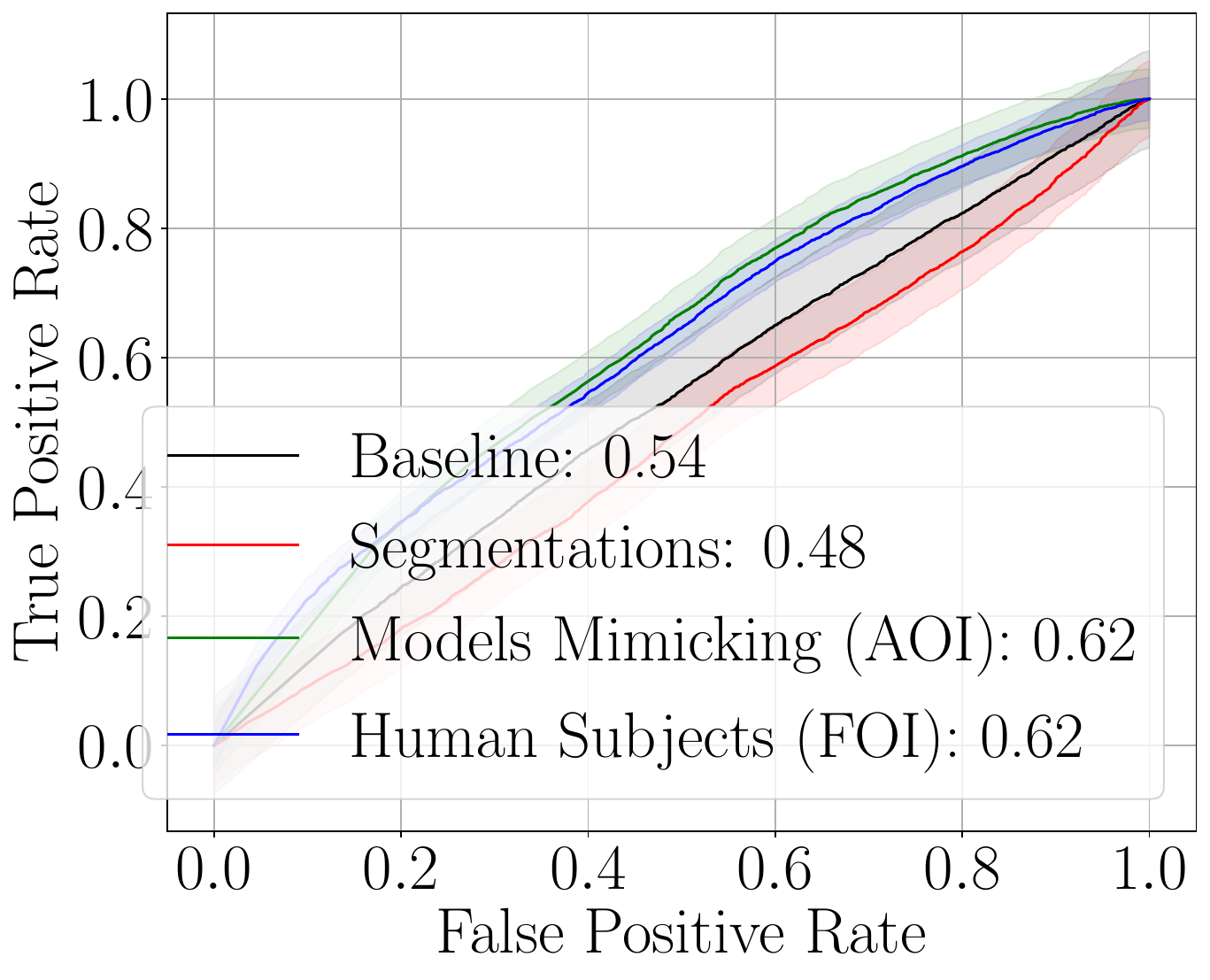}
    \caption{DenseNet}
  \end{subfigure}
  \hfill
  \begin{subfigure}[t]{.33\textwidth}
    \centering
    \includegraphics[width=\linewidth]{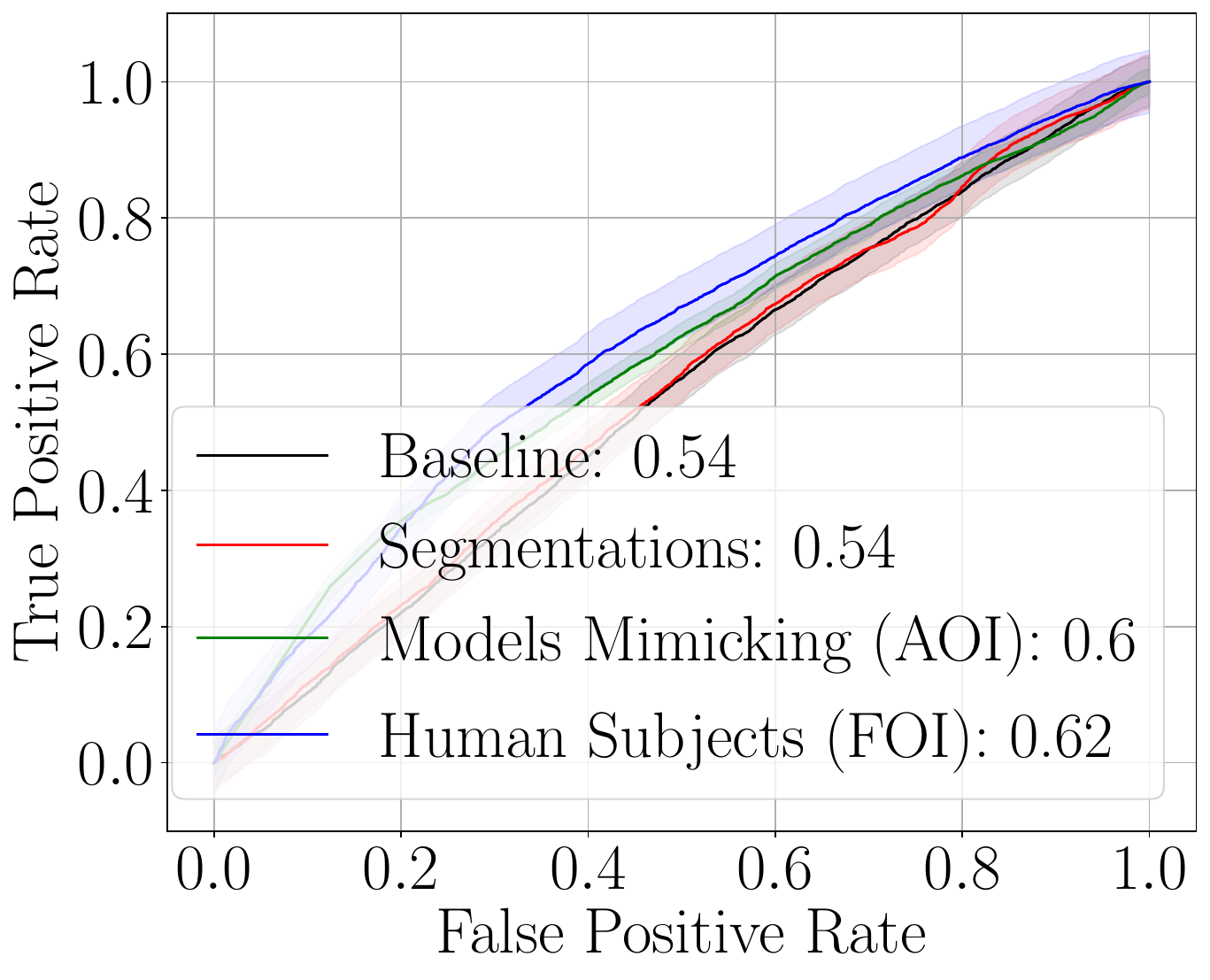}
    \caption{Inception}
  \end{subfigure}
    \caption{Same as in Fig. \ref{fig:results-AUC-iris}, except for {\bf synthetic face detection}.}
    \label{fig:results-AUC-face}
\end{figure*}

%% file: main.bbl
\begin{thebibliography}{10}\itemsep=-1pt

\bibitem{casia-database}
{Chinese Academy of Sciences Institute of Automation}.
\newblock http://www.cbsr.ia.ac.cn/china/Iris\%20Databases\%20CH.asp.
\newblock Accessed: 03-12-2021.

\bibitem{Banerjee_IJCB_2017}
S.~Banerjee, J.~S. Bernhard, W.~J. Scheirer, K.~W. Bowyer, and P.~J. Flynn.
\newblock {SREFI: Synthesis of realistic example face images}.
\newblock In {\em IEEE International Joint Conference on Biometrics (IJCB)}, pages 37--45, 2017.

\bibitem{boyd2022human}
A.~Boyd, K.~W. Bowyer, and A.~Czajka.
\newblock Human-aided saliency maps improve generalization of deep learning.
\newblock In {\em Proceedings of the IEEE/CVF Winter Conference on Applications of Computer Vision, Waikoloa, HI, USA}, pages 2735--2744, 2022.

\bibitem{boyd2020iris}
A.~Boyd, Z.~Fang, A.~Czajka, and K.~W. Bowyer.
\newblock Iris presentation attack detection: Where are we now?
\newblock {\em Pattern Recognition Letters}, 138:483--489, 2020.

\bibitem{boyd2021cyborg}
A.~Boyd, P.~Tinsley, K.~Bowyer, and A.~Czajka.
\newblock Cyborg: Blending human saliency into the loss improves deep learning-based synthetic face detection.
\newblock In {\em 2023 IEEE/CVF Winter Conference on Applications of Computer Vision (WACV)}, pages 6097--6106, 2023.

\bibitem{casula2021livdet}
R.~Casula, M.~Micheletto, G.~Orr{\`u}, R.~Delussu, S.~Concas, A.~Panzino, and G.~L. Marcialis.
\newblock Livdet 2021 fingerprint liveness detection competition-into the unknown.
\newblock In {\em 2021 IEEE international joint conference on biometrics (IJCB)}, pages 1--6. IEEE, 2021.

\bibitem{stargan}
Y.~Choi, Y.~Uh, J.~Yoo, and J.-W. Ha.
\newblock Stargan v2: Diverse image synthesis for multiple domains.
\newblock In {\em Proceedings of the IEEE/CVF conference on computer vision and pattern recognition}, pages 8188--8197, 2020.

\bibitem{crum2023seriously}
C.~R. Crum and C.~Coglianese.
\newblock Taking training seriously: Human guidance and management-based regulation of artificial intelligence.
\newblock {\em arXiv preprint}, 2024.

\bibitem{crum2023explain}
C.~R. Crum, P.~Tinsley, A.~Boyd, J.~Piland, C.~Sweet, T.~Kelley, K.~Bowyer, and A.~Czajka.
\newblock Explain to me: Salience-based explainability for synthetic face detection models.
\newblock {\em IEEE Transactions on Artificial Intelligence}, 2023.

\bibitem{czajka2023irisrepo}
A.~Czajka.
\newblock Iris recognition designed for post-mortem and diseased eyes.
\newblock \url{https://github.com/aczajka/iris-recognition---pm-diseased-human-driven-bsif}, 2023.

\bibitem{das2020iris}
P.~Das, J.~McFiratht, Z.~Fang, A.~Boyd, G.~Jang, A.~Mohammadi, S.~Purnapatra, D.~Yambay, S.~Marcel, M.~Trokielewicz, et~al.
\newblock Iris liveness detection competition (livdet-iris)-the 2020 edition.
\newblock In {\em 2020 IEEE international joint conference on biometrics (IJCB)}, pages 1--9. IEEE, 2020.

\bibitem{Das_IJCB_2020}
P.~{Das}, J.~{Mcfiratht}, Z.~{Fang}, A.~{Boyd}, G.~{Jang}, A.~{Mohammadi}, S.~{Purnapatra}, D.~{Yambay}, S.~{Marcel}, M.~{Trokielewicz}, P.~{Maciejewicz}, K.~{Bowyer}, A.~{Czajka}, S.~{Schuckers}, J.~{Tapia}, S.~{Gonzalez}, M.~{Fang}, N.~{Damer}, F.~{Boutros}, A.~{Kuijper}, R.~{Sharma}, C.~{Chen}, and A.~{Ross}.
\newblock {Iris Liveness Detection Competition (LivDet-Iris) - The 2020 Edition}.
\newblock In {\em 2020 IEEE International Joint Conference on Biometrics (IJCB)}, pages 1--9, 2020.

\bibitem{ImageNet}
J.~Deng, W.~Dong, R.~Socher, L.-J. Li, K.~Li, and L.~Fei-Fei.
\newblock Imagenet: A large-scale hierarchical image database.
\newblock In {\em 2009 IEEE Conference on Computer Vision and Pattern Recognition}, pages 248--255, 2009.

\bibitem{Galbally_ICB_2012}
J.~Galbally, J.~Ortiz-Lopez, J.~Fierrez, and J.~Ortega-Garcia.
\newblock Iris liveness detection based on quality related features.
\newblock In {\em 2012 5th IAPR Int. Conf. on Biometrics (ICB)}, pages 271--276, New Delhi, India, March 2012. IEEE.

\bibitem{he2016deep}
K.~He, X.~Zhang, S.~Ren, and J.~Sun.
\newblock Deep residual learning for image recognition.
\newblock In {\em 2016 IEEE Conference on Computer Vision and Pattern Recognition (CVPR)}, pages 770--778, 2016.

\bibitem{huang2017densely}
G.~Huang, Z.~Liu, L.~Van Der~Maaten, and K.~Q. Weinberger.
\newblock Densely connected convolutional networks.
\newblock In {\em Proceedings of the IEEE conference on computer vision and pattern recognition}, pages 4700--4708, 2017.

\bibitem{Iakubovskii:2019}
P.~Iakubovskii.
\newblock Segmentation models pytorch.
\newblock \url{https://github.com/qubvel/segmentation_models.pytorch}, 2019.

\bibitem{karras2017progressive}
T.~Karras, T.~Aila, S.~Laine, and J.~Lehtinen.
\newblock {Progressive Growing of GANs for Improved Quality, Stability, and Variation}.
\newblock {\em arXiv preprint arXiv:1710.10196}, 2017.

\bibitem{Karras2020ada}
T.~Karras, M.~Aittala, J.~Hellsten, S.~Laine, J.~Lehtinen, and T.~Aila.
\newblock Training generative adversarial networks with limited data.
\newblock In {\em Proc. NeurIPS}, 2020.

\bibitem{karras2021sg3}
T.~Karras, M.~Aittala, S.~Laine, E.~H\"ark\"onen, J.~Hellsten, J.~Lehtinen, and T.~Aila.
\newblock Alias-free generative adversarial networks.
\newblock {\em Proc. NeurIPS}, 2021.

\bibitem{StyleGAN2}
T.~Karras, S.~Laine, M.~Aittala, J.~Hellsten, J.~Lehtinen, and T.~Aila.
\newblock Analyzing and improving the image quality of stylegan.
\newblock In {\em Proceedings of the IEEE/CVF conference on computer vision and pattern recognition}, pages 8110--8119, 2020.

\bibitem{Kohli_ICB_2013}
N.~Kohli, D.~Yadav, M.~Vatsa, and R.~Singh.
\newblock Revisiting iris recognition with color cosmetic contact lenses.
\newblock In {\em {IEEE} Int. Conf. on Biometrics (ICB)}, pages 1--7, Madrid, Spain, June 2013. IEEE.

\bibitem{Kohli_BTAS_2016}
N.~Kohli, D.~Yadav, M.~Vatsa, R.~Singh, and A.~Noore.
\newblock Detecting medley of iris spoofing attacks using desist.
\newblock In {\em {IEEE} Int. Conf. on Biometrics: Theory Applications and Systems (BTAS)}, pages 1--6, Niagara Falls, NY, USA, Sept 2016. IEEE.

\bibitem{Sung_OE_2007}
S.~J. Lee, K.~R. Park, Y.~J. Lee, K.~Bae, and J.~H. Kim.
\newblock {Multifeature-based fake iris detection method}.
\newblock {\em Optical Engineering}, 46(12):1 -- 10, 2007.

\bibitem{liaw2018tune}
R.~Liaw, E.~Liang, R.~Nishihara, P.~Moritz, J.~E. Gonzalez, and I.~Stoica.
\newblock Tune: A research platform for distributed model selection and training.
\newblock {\em arXiv preprint arXiv:1807.05118}, 2018.

\bibitem{linsley2018learning}
D.~Linsley, D.~Shiebler, S.~Eberhardt, and T.~Serre.
\newblock Learning what and where to attend.
\newblock {\em arXiv preprint arXiv:1805.08819}, 2018.

\bibitem{mishra2021image}
S.~Mishra, D.~Z. Chen, and X.~S. Hu.
\newblock Image complexity guided network compression for biomedical image segmentation.
\newblock {\em ACM Journal on Emerging Technologies in Computing Systems (JETC)}, 18(2):1--23, 2021.

\bibitem{orru2019livdet}
G.~Orr{\`u}, R.~Casula, P.~Tuveri, C.~Bazzoni, G.~Dessalvi, M.~Micheletto, L.~Ghiani, and G.~L. Marcialis.
\newblock Livdet in action-fingerprint liveness detection competition 2019.
\newblock In {\em 2019 international conference on biometrics (ICB)}, pages 1--6. IEEE, 2019.

\bibitem{Phillips_IVC_2017}
P.~J. Phillips, P.~J. Flynn, and K.~W. Bowyer.
\newblock Lessons from collecting a million biometric samples.
\newblock {\em Image and Vision Computing}, 58:96--107, 2017.

\bibitem{piland2023model}
J.~Piland, A.~Czajka, and C.~Sweet.
\newblock Model focus improves performance of deep learning-based synthetic face detectors.
\newblock {\em IEEE Access}, 2023.

\bibitem{purnapatra2021face}
S.~Purnapatra, N.~Smalt, K.~Bahmani, P.~Das, D.~Yambay, A.~Mohammadi, A.~George, T.~Bourlai, S.~Marcel, S.~Schuckers, et~al.
\newblock Face liveness detection competition (livdet-face)-2021.
\newblock In {\em 2021 IEEE International Joint Conference on Biometrics (IJCB)}, pages 1--10. IEEE, 2021.

\bibitem{ronneberger2015u}
O.~Ronneberger, P.~Fischer, and T.~Brox.
\newblock U-net: Convolutional networks for biomedical image segmentation.
\newblock In {\em Medical Image Computing and Computer-Assisted Intervention--MICCAI 2015: 18th International Conference, Munich, Germany, October 5-9, 2015, Proceedings, Part III 18}, pages 234--241. Springer, 2015.

\bibitem{szegedy2016inceptionv4}
C.~Szegedy, S.~Ioffe, V.~Vanhoucke, and A.~Alemi.
\newblock Inception-v4, inception-resnet and the impact of residual connections on learning, 2016.

\bibitem{szegedy2016rethinking}
C.~Szegedy, V.~Vanhoucke, S.~Ioffe, J.~Shlens, and Z.~Wojna.
\newblock Rethinking the inception architecture for computer vision.
\newblock In {\em Proceedings of the IEEE conference on computer vision and pattern recognition}, pages 2818--2826, 2016.

\bibitem{tinsley2023iris}
P.~Tinsley, S.~Purnapatra, M.~Mitcheff, A.~Boyd, C.~Crum, K.~Bowyer, P.~Flynn, S.~Schuckers, A.~Czajka, M.~Fang, et~al.
\newblock Iris liveness detection competition (livdet-iris)--the 2023 edition.
\newblock In {\em 2023 IEEE International Joint Conference on Biometrics (IJCB)}, pages 1--10. IEEE, 2023.

\bibitem{Trokielewicz_BTAS_2015}
M.~{Trokielewicz}, A.~{Czajka}, and P.~{Maciejewicz}.
\newblock Assessment of iris recognition reliability for eyes affected by ocular pathologies.
\newblock In {\em {IEEE} Int. Conf. on Biometrics: Theory Applications and Systems (BTAS)}, pages 1--6, 2015.

\bibitem{Trokielewicz_IVC_2020}
M.~Trokielewicz, A.~Czajka, and P.~Maciejewicz.
\newblock Post-mortem iris recognition with deep-learning-based image segmentation.
\newblock {\em Image and Vision Computing}, 94:103866, 2020.

\bibitem{van2023probabilistic}
T.~van Sonsbeek, X.~Zhen, D.~Mahapatra, and M.~Worring.
\newblock Probabilistic integration of object level annotations in chest x-ray classification.
\newblock In {\em Proceedings of the IEEE/CVF Winter Conference on Applications of Computer Vision}, pages 3630--3640, 2023.

\bibitem{wang2024gazegnn}
B.~Wang, H.~Pan, A.~Aboah, Z.~Zhang, E.~Keles, D.~Torigian, B.~Turkbey, E.~Krupinski, J.~Udupa, and U.~Bagci.
\newblock Gazegnn: A gaze-guided graph neural network for chest x-ray classification.
\newblock In {\em Proceedings of the IEEE/CVF Winter Conference on Applications of Computer Vision}, pages 2194--2203, 2024.

\bibitem{Wei_ICPR_2008}
Z.~Wei, T.~Tan, and Z.~Sun.
\newblock Synthesis of large realistic iris databases using patch-based sampling.
\newblock In {\em Int. Conf. on Pattern Recognition (ICPR)}, pages 1--4, Tampa, FL, USA, Dec 2008. IEEE.

\bibitem{yambay2017livdet}
D.~{Yambay}, B.~{Becker}, N.~{Kohli}, D.~{Yadav}, A.~{Czajka}, K.~W. {Bowyer}, S.~{Schuckers}, R.~{Singh}, M.~{Vatsa}, A.~{Noore}, D.~{Gragnaniello}, C.~{Sansone}, L.~{Verdoliva}, L.~{He}, Y.~{Ru}, H.~{Li}, N.~{Liu}, Z.~{Sun}, and T.~{Tan}.
\newblock {LivDet-Iris 2017 -- Iris Liveness Detection Competition 2017}.
\newblock In {\em IEEE International Joint Conference on Biometrics (IJCB)}, pages 733--741, 2017.

\bibitem{Yambay_IJCB_2017}
D.~Yambay, B.~Becker, N.~Kohli, D.~Yadav, A.~Czajka, K.~W. Bowyer, S.~Schuckers, R.~Singh, M.~Vatsa, A.~Noore, D.~Gragnaniello, C.~Sansone, L.~Verdoliva, L.~He, Y.~Ru, H.~Li, N.~Liu, Z.~Sun, and T.~Tan.
\newblock {LivDet Iris 2017 -- Iris Liveness Detection Competition 2017}.
\newblock In {\em {IEEE} Int. Joint Conf. on Biometrics (IJCB)}, pages 1--6, Denver, CO, USA, 2017. IEEE.

\bibitem{yambay2018livdet}
D.~Yambay, S.~Schuckers, S.~Denning, C.~Sandmann, A.~Bachurinski, and J.~Hogan.
\newblock Livdet 2017-fingerprint systems liveness detection competition.
\newblock In {\em 2018 IEEE 9th international conference on biometrics theory, applications and systems (BTAS)}, pages 1--9. IEEE, 2018.

\bibitem{Yambay_ISBA_2017}
D.~Yambay, B.~Walczak, S.~Schuckers, and A.~Czajka.
\newblock Livdet-iris 2015 - iris liveness detection competition 2015.
\newblock In {\em {IEEE} Int. Conf. on Identity, Security and Behavior Analysis (ISBA)}, pages 1--6, New Delhi, India, Feb 2017. IEEE.

\bibitem{yu2018bisenet}
C.~Yu, J.~Wang, C.~Peng, C.~Gao, G.~Yu, and N.~Sang.
\newblock Bisenet: Bilateral segmentation network for real-time semantic segmentation.
\newblock In {\em Proceedings of the European conference on computer vision (ECCV)}, pages 325--341, 2018.

\bibitem{zllrunning2019parse}
zllrunning.
\newblock face-parsing.pytorch.
\newblock \url{https://github.com/zllrunning/face-parsing.PyTorch}, 2019.

\end{thebibliography}
